\documentclass[letterpaper, 10 pt, conference]{ieeeconf}  % Comment this line out if you need a4paper

\IEEEoverridecommandlockouts                              % This command is only needed if 
\usepackage{graphicx}
\usepackage{amsmath}
\title{\LARGE \bf
Cognitive Dynamic Systems: A Technical Review of Cognitive Radar  
}

\author{Krishanth Krishnan$^{ 1,*}$, Taralyn Schwering$^{ 1,*}$ and Saman Sarraf$^{ 1,*}$
\thanks{$^{1}$Krishanth Krishnan, Taralyn Schwering and Saman Sarraf were with the Department of Electrical and Computer Engineering, McMaster University, Hamilton, ON, L8S 4L8, Canada 
		{Corresponding Author: \tt\small samansarraf@ieee.org}}%
\thanks{$^{*}$All authors equally contributed to this work.}%
}

\begin{document}

\maketitle
\thispagestyle{empty}
\pagestyle{empty}

%%%%%%%%%%%%%%%%%%%%%%%%%%%%%%%%%%%%%%%%%%%%%%%%%%%%%%%%%%%%%%%%%%%%%%%%%%%%%%%%
\begin{abstract}

In this report we discuss the mechanisms of cognitive radar. This report is organized as follows:
We start with the history of cognitive radar, where origins of the PAC, Fuster?s research on cognition and principals of cognition are provided. Fuster describes five cognitive functions: perception, memory, attention, language, and intelligence. We describe the Perception-Action Cyclec (PAC) as it applies to cognitive radar, and then discuss long-term memory, memory storage, memory retrieval and working memory. A comparison between memory in human cognition and cognitive radar is given as well. Attention is another function described by Fuster, and we have given the comparison of attention in human cognition and cognitive radar. We talk about the four functional blocks from the PAC: Bayesian filter, feedback information, dynamic programming and state-space model for the radar environment. Then, to show that the PAC improves the tracking accuracy of Cognitive Radar over Traditional Active Radar, we have provided simulation results. In the simulation, three nonlinear filters:  Cubature Kalman Filter (CKF), Unscented Kalman Filter (UKF) and Extended Kalman Filter (EKF) are compared. Based on the results, radars implemented with CKF perform better than the radars implemented with UKF or radars implemented with EKF. Further, radar with EKF has the worst accuracy and has the biggest computation load because of derivation and evaluation of Jacobian matrices. In addition, we suggest using the concept of risk management to better control parameters and improve performance in cognitive radar, especially in noise reduction.  We believe, spectrum sensing can be seen as a potential interest to be used in cognitive radar and we propose a new approach Probabilistic Independent Component Analysis (PICA) which will presumably reduce noise based on estimation error in cognitive radar. Parallel computing is a concept based on divide and conquers mechanism, and in this report we suggest to use the parallel computing approach in cognitive radar by doing complicated calculations or tasks to reduce processing time, increase speed of system and higher precision. With the increased use of parallel computing in other fields of engineering, General-purpose computing on graphics processing units (GPGPU) would be the best way to implement parallel computing in cognitive radar.

\end{abstract}

%%%%%%%%%%%%%%%%%%%%%%%%%%%%%%%%%%%%%%%%%%%%%%%%%%%%%%%%%%%%%%%%%%%%%%%%%%%%%%%%
\section{INTRODUCTION AND HISTORY}

\subsection{Fuster and the origins of the Perception-Action Cycle}
Human Cognition refers to high level brain functions that deal with understanding information, processing it and retaining knowledge. Cognition in particular deals with the logical or analytical aspects as opposed to the emotional ones. From Heraclitos  in the 5th century BC to contemporary psychophysicists, neuroscientists and psychologists have always believed that perception is the representation of the world which is entering the mind through the senses. In modern neuroscience, perception is widely assumed to be reducible to the effects of sensory stimuli upon dedicated receptors, pathways and nerve cells \cite{fuster2003cortex}.
The perception action cycle has its origins in the 1920?s when Jacob von Uexküll described a sensory-motor cycle observed in the behaviour of primitive animals. He observed that an organism will perceive their environment through their senses and base their actions on theses perceptions. Joaquin Fuster later extended this idea to higher animals by describing the perception-action cycle (PAC). As described by Fuster, the perception action cycle is more complex and is based in neurobiology rather than the behavioural observation done by Uexküll. The cycle, which determines an animal?s actions, begins when the animal ?perceives? its immediate environment through its senses. The act of perception includes recognition and updating memory, with the goal of building a representation of the world. This process takes place in the posterior cortex of the brain. Next, the animal?s behaviour or selection of actions is based on the perceptions of the environment that the animal has built. These actions then act on and change the environment and therefore change the animals subsequent perception of its environment. The cycle is completed with these new perceptions and the actions based off them.
\subsection{Cognitive Functions}
In his book, \textbf{CORTEX AND MIND: UNIFYING COGNITION}, Fuster describes five cognitive functions: perception, memory, attention, language, and intelligence.
Perception relates to one?s understanding of the world or environment around them. It goes beyond simple sensory inputs and includes functions of memory and attention. In neurobiological terms, perception involves the activation of cognits, or cortical networks, that make up memories.
Memory deals with the capacity to retain information about oneself and one?s environment. This information includes both conscious and unconscious knowledge, the mental traces of experience, past events, learned facts and relationships between facts [1]. The formation of memory (sometimes referred to as memory encoding) involves the formation of cortical networks and the strengthening of synapses that form these networks to create associations by which memories can be retrieved. From an information-processing point of view, the function of memory is to encode and to store the received signal and then recall this information when required [2]. 
Attention refers to the efficient use of resources when dealing with a task. It is displayed in the selective allocation of neural resources to that task through excitation and inhibition of cortical networks. The role of attention is to select one of the cortical networks at a time and to keep it active for as long as it serves a cognitive function or the achievement of a behavioral goal [1] [3]. Unlike memory and perception, there is no separate structure for attention as it is evident in processes rather than localized areas of the brain. From the engineering point of view, the attention provides for the effective, efficient and well-organized utilization of computational resources in a cognitive dynamic system, so as to avoid the information overload problem [2].
While it is not very difficult to define language as a set of rules for communication, it is not necessarily as clear how it fits in with the other four functions of cognition. Essentially, the mechanism of language is localized in perceptual and executive areas of the cortex [1]. Perceptual areas are responsible for understanding incoming speech or written words. On the other hand executive areas are responsible for forming transforming complex ideas into understandable outgoing speech. 
Intelligence is the most complex of the five functions of cognition as it is distributed throughout the other four functions [1]. Intelligence is the ability of a cognitive dynamic system to continually adjust itself through an adaptive process by making the PAC respond to new changes in the environment so as to create new forms of action and behavior in the actuator [3]. Specifically, the function of intelligence is to enable a decision-making algorithm to select the optimal solution [2].
\subsection{Cognitive Dynamic systems}
The first papers written on cognitive dynamic systems in 2005 and 2006 did not reference Fuster, the perception action cycle or any of the principals of cognition. Cognitive radar however was specifically inspired by echolocation in bats and it is not hard to see how the perception-action cycle applies directly to bats? ability to navigate their surroundings and locate food based on the sounds they emit.
Later papers on the topic of cognitive dynamic systems have incorporated ideas drawn from Fuster?s work. Due to the similarity of structure between cognitive radar and human cognition, specifically the overriding structure of the perception-action cycle, cognitive radar is well poised to benefit from Fuster?s work on the functions of cognition.

\section{BACKGROUND AND ALGORITHMS}
\subsection{Perception Action Cycle}
\begin{figure}[h!]
	\includegraphics[width=\linewidth]{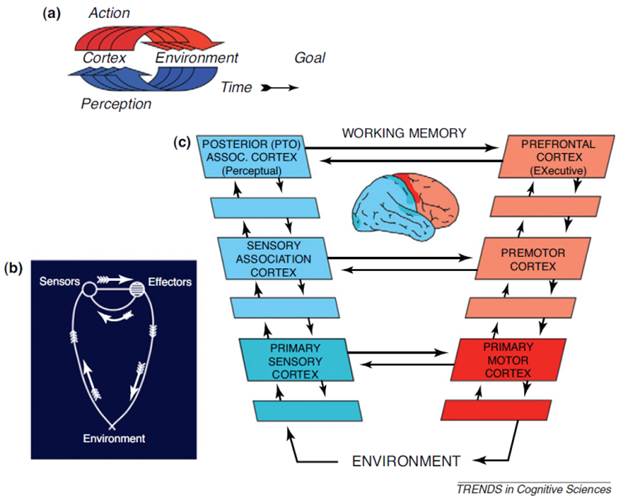}
	\caption{Schematic structure of the cortical perception-action cycle around a diagram of the human brain [4] (a) Basic diagram of the PAC toward a goal through cortex and environment, (b) Basic diagram of the sensory-motor cycle from Von Uexküll, with the internal nervous feedback from efferent to sensors, (c) General view of the connective framework of the PAC in the primate cortex.}
	\label{Fig.1}
\end{figure}

The function of the perception-action cycle (PAC) is to generate information by receiving and processing signals from the environment. Thus, a cognitive dynamic system, for example cognitive radar must be equipped with an appropriate set of sensors to receive signals and learn from the environment [3]. The perception-action cycle is described in \ref{Fig.1}. 
\subsection{Memory}
\subsubsection{Long-term Memory}
The executive and perceptual memories within cognitive radar are analogous to long term memory in the human (or animal) brain. In the human brain, long term memory is responsible for storing information so that it can be recalled and used by working memory. Working memory in cognitive radar and the human brain will be discussed later. 
According to Fuster, stored memories consist of cortical networks or cognits, which contain information within their structural mesh [1]. The bits of information are interconnected and organised through associations which allow us to recall old memories and store new memories efficiently. Most functions related to long-term memory in both humans and cognitive radar fall under one of two concepts: encoding (or storing of information in these cortical networks) and retrieving the information for use at a later time.
\subsubsection{Encoding}
In [5] a working definition for cognitive dynamic systems is proposed as systems that ?build up rules of behavior over time through learning from continuous experiential interactions with the environment, and thereby deal with environmental uncertainties.? This process of learning is analogous to what some neuroscientists call encoding of memories. Physiologically, Fuster refers to this encoding as the building and strengthening of synapses or neural connections between cognits through Hebbian learning [1]. Hebbian learning occurs when 2 cells or 2 groups of cells are consistently activated at the same time. These cells will become associated such that they are activated together in the future. This encoding process is constantly being reinforced through repeated action. In other words, our memories of certain facts or motor processes are improved upon experience. This idea is central in cognitive radar as well, in that the radar system with memory displays improved efficiency with each iteration of the perception-action cycle [3].
The formation of memories (or knowledge) in a cognitive radar system happens through machine learning [6]. Through the use of some training set, a set of rules is built in the perceptual memory to properly predict (match, recognize) the model of the environment (system-model space) based on sensory input information (measurement space). In the executive memory, the rules are to properly select the ideal output waveform (from the transmit-waveform library) based on feedback information (indirect view of the environment supplied through the receiver).
This initial learning happens through unsupervised learning using a replicator neural network. Replicator neural network consists of an encoder, which determines rules for matching input to abstract features, and a decoder, which does the opposite. By coupling them together and comparing the output of the decoder to the original training set, we can use the backpropagation algorithm to pass this error back through the layers on the network and train it. When this initial training phase is complete, the decoder part is removed as it is no longer needed and just the encoder (to match input to abstract features) is used for each the perceptual and the executive memories [6].
At the output of the encoder we have a set of abstract features derived from the original input.  In perceptual memory, this output must be matched to a signal in the system-modal library. Rules for this matching process are determined through supervised learning where the ideal signal in the system-model library is known for any given input of radar returns. A similar process is used in the executive memory as well, where the ideal transmit-waveform is known for any given feedback input.
\subsubsection{Retrieval}
The associative structure of memory enables efficient recall or recognition of cognits in addition to efficient storage. On the perceptual side, relevant cognits are activated by related sensory inputs through associations. On the executive side, cognits are activated by feedback signals from the perceptual memory.
In addition to being arranged associatively, both perceptual and executive memories in the perception-action cycle are arranged hierarchically in the human brain. This means that inputs are filtered through several layers, each layer extracting more and more specific information about the features of the input. In both the perceptual and executive areas of the human brain, the lowest levels of the hierarchy, known a phyletic memory, act as the foundation of memory off of which all other memory is based [1]. In the posterior cortex (associated with perception) elementary sensory features are detected while basic motor functions are integrated in the prefrontal cortex (associated with execution). As you go up the memory hierarchy, memory gets more complex and abstract. The top layer of perceptual memory, semantic memory, has to do with understanding of complex situations and concept while in executive memory the top layer has to do with planning and rules to achieve specific goals [1]. 
Another important feature of hierarchical memory is that connections exist between perceptual and executive memory at all levels. Connections at lower levels of the hierarchy take care of automatic and routine activities while more complex perceptions and actions involve higher layers of memory.
In cognitive radar hierarchical memory is useful because it allows us to reduce a complex problem into simpler intermediary steps. The first layer of perceptual memory extracts the broad features from the sensory input while the subsequent layers extract finer and finer features of these initial features with the goal of simplifying the input in order to match it to an appropriate system-model. Hierarchical memory in executive side works in a similar way, using feedback information from the receiver as input and building up features of the output waveform in order to best illuminate the environment. 
\subsubsection{Working memory}
In humans and animals, working memory refers to the activation of cognits held over from perception to execution [1]. It is sometimes referred to as short-term memory since it has a limited capacity so information only stays in the working memory for a short period of time (while one is actively thinking about a piece of knowledge). Because of this limited capacity, the cognits activated in the working memory depends highly on attention. Attention will be discussed in further detail in the next section. 
The purpose of working memory in cognitive radar is to couple the perceptual and working memories together and to activate the appropriate output in the executive memory based on the input from the perceptual memory [2]. In the human brain, working memory is able to sustain the activation of perceptual cognits after sensory stimulus has disappeared [1]. This allows the input to the executive memory to remain while action is being determined (planned, executed). This function of working memory is useful as well in cognitive radar. The working memory is able to hold the feedback information from the receiver indefinitely while the transmitter selects a transmit waveform. It can also continue to hold this information if there is a gap in the signals being received at the transmitter.
In addition to holding onto the feedback information, working memory is able to predict the consequences of actions based off information from both sides of memory. The predictions are then fed back into the perceptual and executive memories where they facilitate the learning process [3].
\subsection{Attention}
As mentioned earlier, attention refers to the efficient activation of cognits for a specific task. There is no specific structure responsible for attention in the human brain, rather it is a process or mechanism that is distributed throughout the perception-action cycle and memory structures. In the human brain attention is responsible for activating relevant cognits in memory while inhibiting other cognits. Similarly, attention is used in cognitive radar to localize a search area when matching a set of abstract features to a signal in the system-modal library or in the transmit-waveform library, using the so-called ?explore-exploit? strategy [6]. 
Using the point in the library selected in the previous iteration of the perception-action cycle as a centre point, a small group of points is searched in the matching process rather than an extensive search of the entire library. This strategy reduces the computational resources used in this matching process as well as stabilizing the output of the system [3].
\subsection{Cognitive Radar}
Mostly, there are two functional parts in cognitive radar shown in Error! Reference source not found. which are the perception-action cycle and distributed memory. Perception, which is performed in the receiver, plays two important roles: the detection of the unknown target and its tracking behavior across time [2].
As discussed above, the perception-action cycle generates an information gain when processing signals from radar environment in which the magnitude of the gain increases from the last cycle to the next. The state-estimation step mathematically plays the role of perception in cognitive radar so that the feedback information can be formulated using the state-estimation vector [6].
\begin{figure}[h!]
	\includegraphics[width=\linewidth]{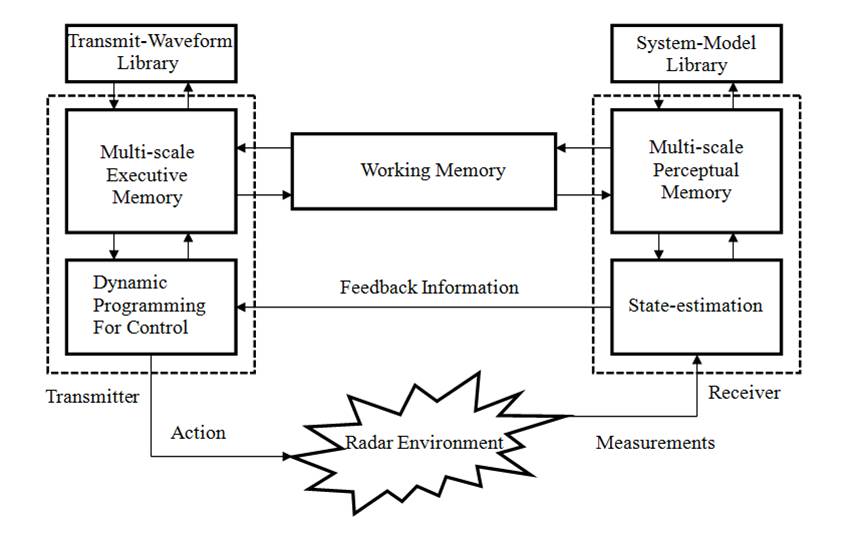}
	\caption{Block Diagram of Cognitive Radar}
	\label{Fig.2}
\end{figure}
Four functional blocks form the basis of the perception-action cycle in cognitive radar: Bayesian filtering, feedback information, dynamic programming and state-space model for the radar environment. These functional blocks are taken care of by the following:
\begin{enumerate}
	\item Bayesian Filtering: this block approximates Bayesian filter for the environmental perception in the receiver. Although this filter is known to be optimal, it is no longer computationally feasible. Therefore, under the Gaussian assumption, the following approaches can be considered:
	\begin{enumerate}
		\item Extended Kalman filter
		\item Unscented Kalman filter
		\item Cubature Kalman filter
	\end{enumerate}
	Under the Gaussian assumption [6], the Bayesian filter reduces to the problem of how to compute moment integrals whose integrands have the following form:
	\begin{equation}\label{Eq.1}
	Nonlinear Function * Gaussian 
	\end{equation}
	\begin{figure}[h!]
		\includegraphics[width=\linewidth]{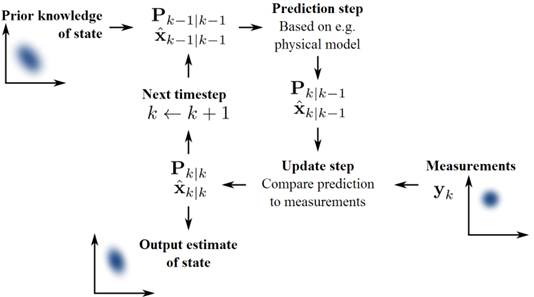}
		\caption{Basic concept of Kalman filtering}
		\label{Fig.3}
	\end{figure}
	\item Shannon's Entropy Versus Fisher Information: The most fundamental and most important notion of information related to random variables is the Kullback-Leibler (KL) information, or equivalently, the Boltzmann entropy, which measure the similarity of two probabilistic models. The Shannon entropy and Fisher?s information matrix are further information quantities, both of which can be directly derived from KL information [8].  Entropy is a measure of the uncertainty about an event in Shannon?s information theory while the expected Fisher information matrix plays an important role in likelihood and Bayes theory, is a local version of KL information. \ref{Fig.4} illustrates the measures of information and their relationship.
	\begin{figure}[h!]
		\includegraphics[width=\linewidth]{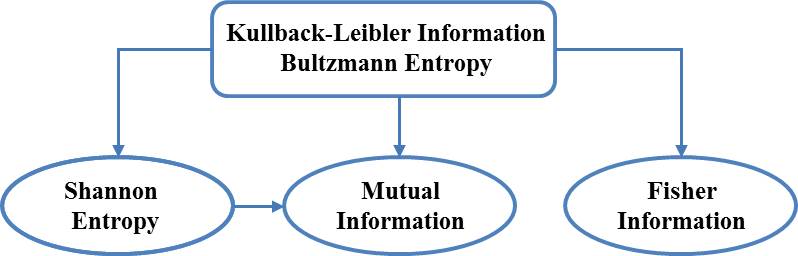}
		\caption{Measures of information and their relationship}
		\label{Fig.4}
	\end{figure}
	Both Shannon entropy (as a global measure of smoothness in the probability density function) and Fisher information (as a local measure of smoothness in the probability density function) can be used in a variational principle to infer the probability density function that describes the phenomenon under consideration. However, the local measure may be preferred in general [6]. 
	\item Posterior Cramér?Rao Lower Bound (PCRLB): In estimation theory, Cramér?Rao Lower Bound (CRLB) presents a lower bound on the variance of estimators of a deterministic parameter. In its simplest form, the bound states that the variance of any unbiased estimator is at least as high as the inverse of the Fisher information. An unbiased estimator which achieves this lower bound is said to be (fully) efficient. Such a solution achieves the lowest possible mean squared error among all unbiased methods, and is therefore the minimum variance unbiased (MVU) estimator [9]. As mentioned previously, the CRLB is a lower bound that represents the lowest possible mean-square error (MSE) in the estimation of deterministic parameters for all unbiased estimators. Computer simulations showed that in cognitive radar, the PCRLB could be reduced to zero which will enable the radar to provide accurate estimates of the estate [6].
	\item Sensitivity Analysis: When testing the robustness of the results of a model or system in the presence of uncertainty, sensitivity analysis can be usefull. We assume that the system and noise on which the Bayesian filter is designed are statistically known. In practice, the true distributions often deviate from the assumed nominal ones thus it would be desirable to modify the filter in order to desensitize modeling errors with respect to implementation approximations. This robustification may deteriorate the performance of the filter. A robust algorithm?s performance must be acceptable for a range of possible deviations from the nominal model. Finally, two approaches are proposed to design filters dealing with the problem of not knowing the true distribution: Stochastic approach and Deterministic approach [6].
	\item Dynamic programming: The function of transmitter in cognitive radar is to control the receiver. In order to have the optimal control on what are received from the environment, we may consider dynamic programming as the method of optimization [3]. 
	
\end{enumerate}

\section{Performance comparison of Nonlinear tracking filters in Cognitive Radar and Traditional Active Radar}
\subsubsection{Objective}
A case study to compare the information gain through use of the perception-action cycle mechanism (first step towards radar cognition) against Traditional Active Radar (TAR) for different nonlinear tracking filters is presented. The perception-action cycle (PAC) mechanism (also referred to in this section as Cognitive Radar) uses selectable transmit waveform while TAR uses fixed transmit waveform. We consider an extensively studied problem in the tracking community: the problem of target reentry in space \cite{haykin2012cognitive}. The target reentry in space problem was considered and simulation results were presented with Cubature Kalman Filter (CKF) in Section 6.12.4 of \cite{haykin2012cognitive}. In this work, we have extended the simulation results to two more nonlinear tracking filters: Extended Kalman Filter (EKF) and Unscented Kalman Filter (UKF).
Simulation parameters used for this case study are given in the appendix.
\subsubsection{Simulation results and comments}
\begin{figure}[h!]
	\includegraphics[width=\linewidth]{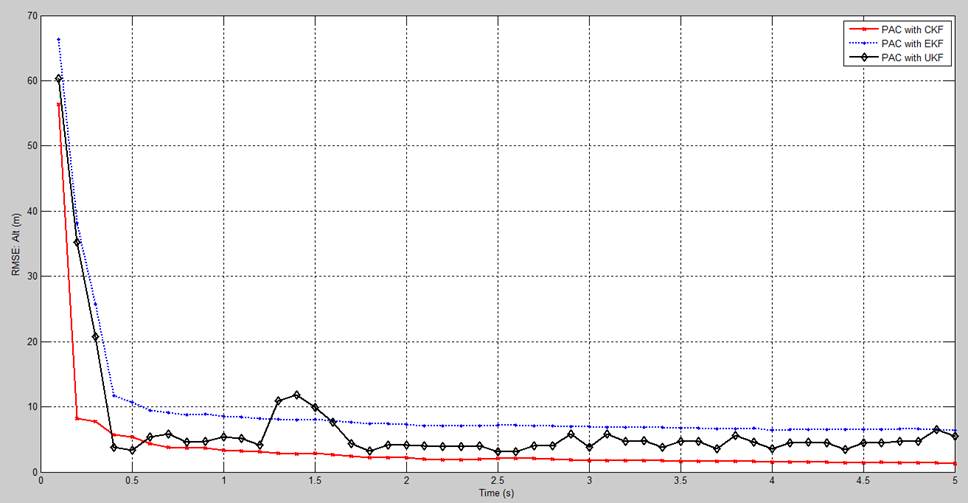}
	\caption{RMSE of altitude for Cognitive Radar with PAC}
	\label{Fig.5}
\end{figure}

\begin{figure}[h!]
	\includegraphics[width=\linewidth]{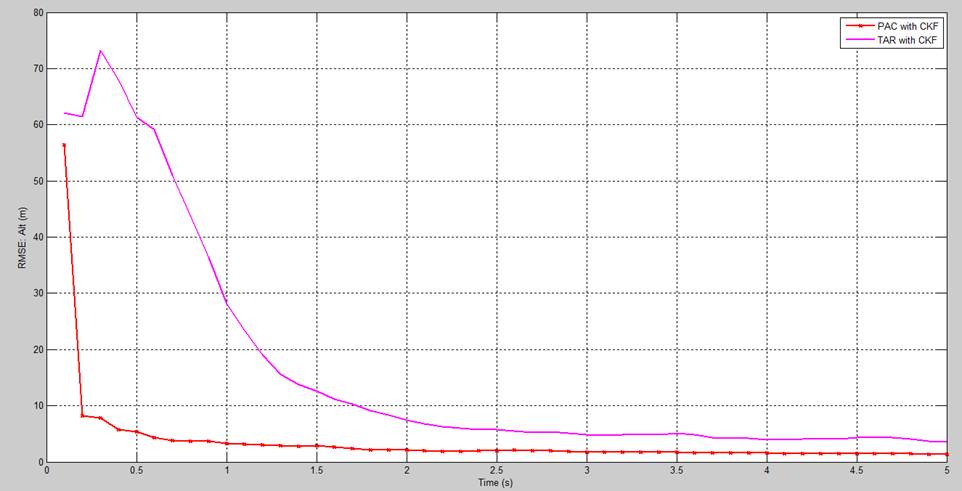}
	\caption{RMSE of altitude for Cognitive Radar and Traditional Active Radar (Both implemented with Cubature Kalman Filter)}
	\label{Fig.6}
\end{figure}

\begin{figure}[h!]
	\includegraphics[width=\linewidth]{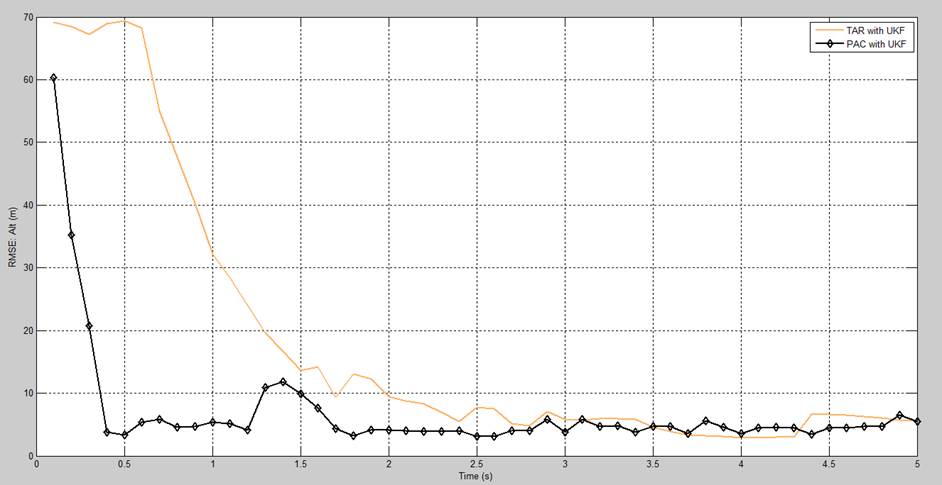}
	\caption{RMSE of altitude for Cognitive Radar and Traditional Active Radar (Both implemented with Unscented Kalman Filter)}
	\label{Fig.7}
\end{figure}

\begin{figure}[h!]
	\includegraphics[width=\linewidth]{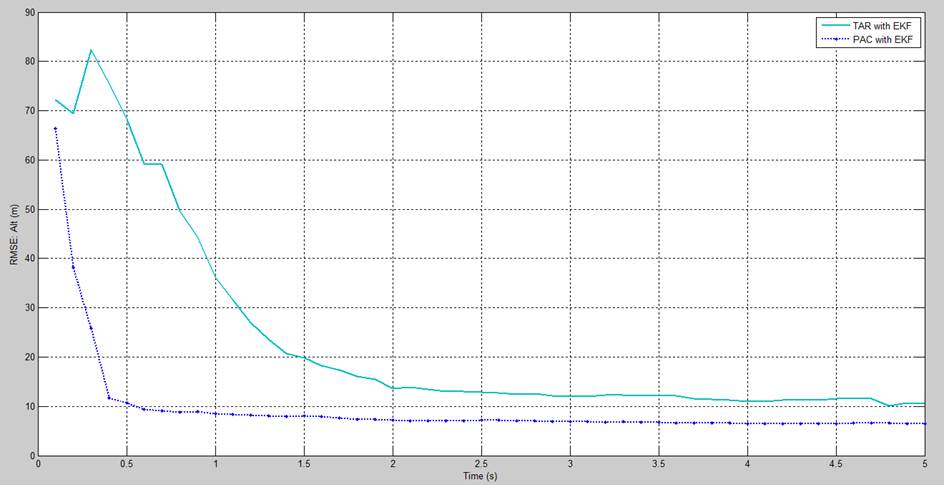}
	\caption{RMSE of altitude for Cognitive Radar and Traditional Active Radar (Both implemented with Extended Kalman Filter)}
	\label{Fig.8}
\end{figure}

\begin{figure}[h!]
	\includegraphics[width=\linewidth]{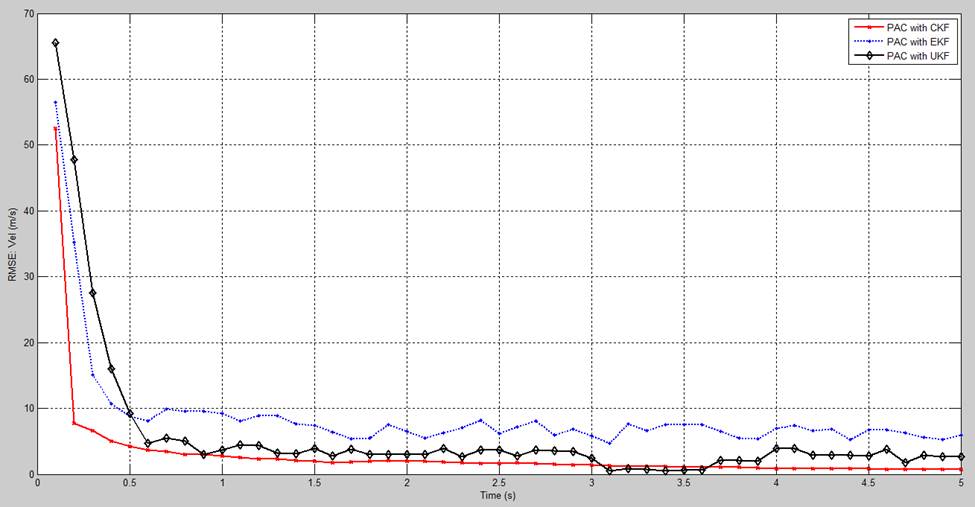}
	\caption{RMSE of Velocity for Cognitive Radar with PAC}
	\label{Fig.9}
\end{figure}

\ref{Fig.5} shows the root-mean-square error (RMSE) of altitude for cognitive radar implemented with three nonlinear filters. It can be seen that the tracking accuracy of EKF is lower compared to the other two filters CKF and UKF. 
\ref{Fig.6} shows RMSE of altitude for Traditional Active Radar and Cognitive Radar (PAC), where both implemented with CKF. Traditional Active Radar uses a fixed transmit-waveform while Cognitive Radar uses selectable transmit-waveform. As expected, Cognitive Radar outperforms the Traditional Active Radar during the whole time period.
\ref{Fig.7} shows RMSE of altitude for Traditional Active Radar and Cognitive Radar, where both radars are implemented with UKF. Similarly \ref{Fig.8} shows the results for both radars implemented with EKF.  In both tests, Cognitive Radar performed better than the Traditional Active Radar.
\ref{Fig.9} shows the RMSE of Velocity for Cognitive Radar implemented with different filters. It can be seen that the tracking accuracy of CKF is better compared to the other two filters, CKF and EKF.
Figure 10 shows RMSE of Velocity for Traditional Active Radar and Cognitive Radar, where both are implemented with CKF. Similarly Figure 11 shows results from both radars implemented with UKF and Figure 12 shows results from both radars implemented with EKF. In the three figures mentioned above, it can be observed that the Cognitive Radar has lower RMSE of velocity compared to the Traditional Active Radar. 
\begin{figure}[h!]
	\includegraphics[width=\linewidth]{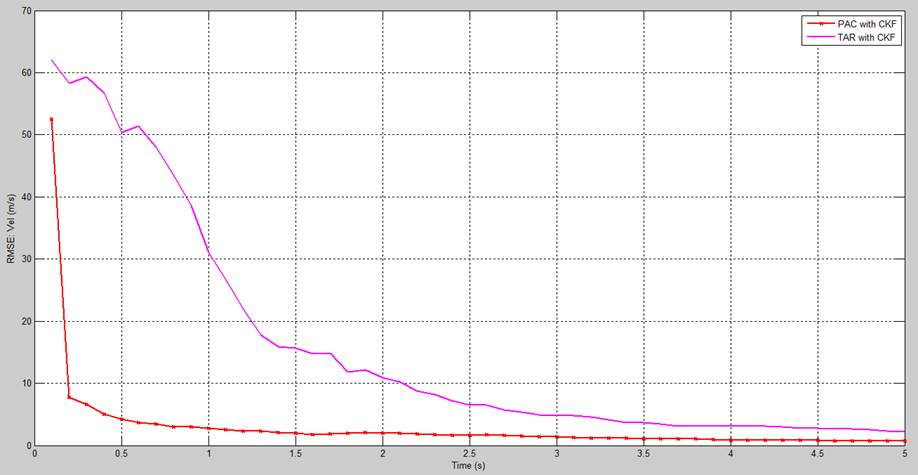}
	\caption{RMSE of Velocity for Cognitive Radar and Traditional Active Radar (Both implemented with Cubature Kalman Filter)}
	\label{Fig.10}
\end{figure}
\begin{figure}[h!]
	\includegraphics[width=\linewidth]{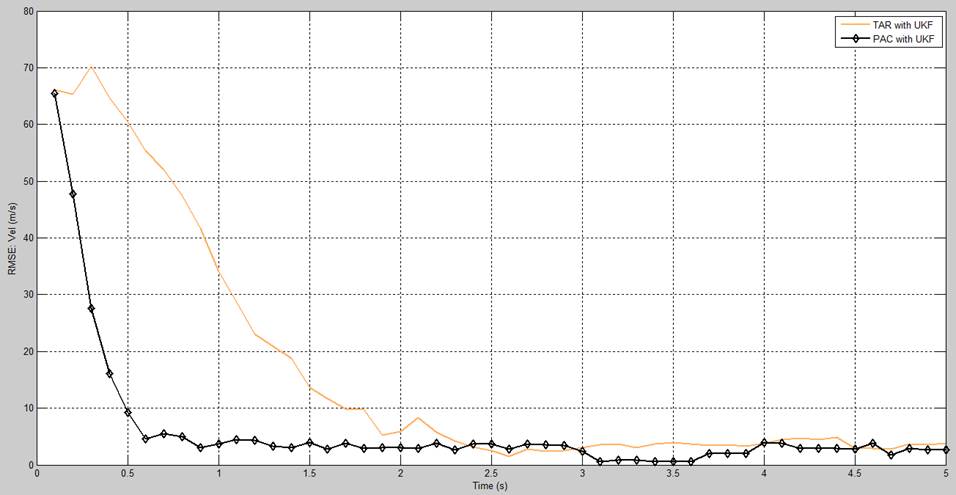}
	\caption{RMSE of Velocity for Cognitive Radar and Traditional Active Radar (Both implemented with Unscented Kalman Filter)}
	\label{Fig.11}
\end{figure}
\begin{figure}[h!]
	\includegraphics[width=\linewidth]{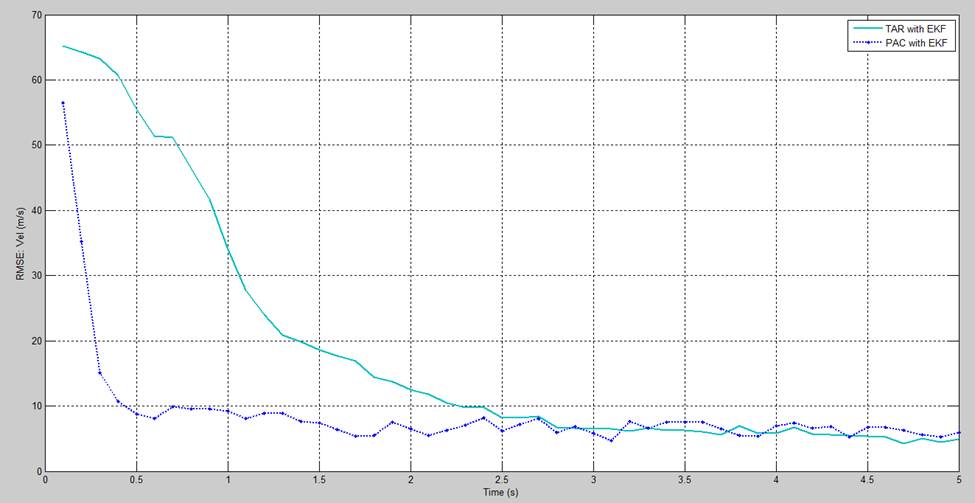}
	\caption{RMSE of Velocity for Cognitive Radar and Traditional Active Radar (Both implemented with Extended Kalman Filter)}
	\label{Fig.12}
\end{figure}
\subsubsection{Simulation Conclusion}

Traditional Active Radar and Cognitive Radar are compared with three nonlinear Kalman filters. RMSE of altitude and RMSE of velocity are used as the performance measure for the comparison purpose. It can be concluded that in all the tests Cognitive Radar performs better than the Traditional Active Radar.  Further, radars implemented with CKF perform better than the radars implemented with UKF or radars implemented with EKF. Radar with EKF has the worst accuracy and has the biggest computation load because of derivation and evaluation of Jacobian matrices. No linearization steps are required for EKF and UKF, so the simulation time is less than the radar implemented with EKF.
\section{System Improvement}
\subsection{Risk Management and Cognitive Radar}

\begin{figure}[h!]
	\includegraphics[width=\linewidth]{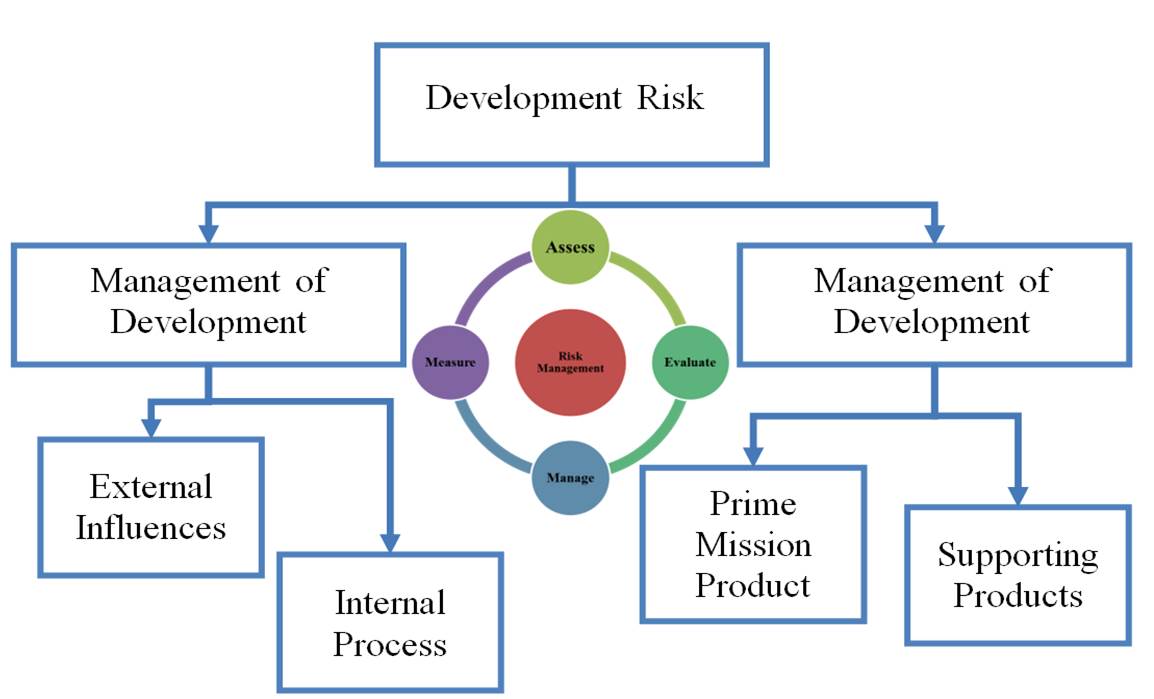}
	\caption{Risk Hierarchy: Assess? Evaluate? Manage? Measure? Assess}
	\label{Fig.13}
\end{figure}

When looking at cognitive radar as a complicated system, there are many parameters and factors to control in order to improve the system performance and minimize any risk which might appear as error. In such case, the concept of risk management coming from economics can be helpful in extracting risk factors from our system and in solving computational problems such as reducing noise. Risk management is an organized method for identifying and measuring risk and for selecting, developing, and implementing options for the handling of risk \cite{leonard1999systems} . \ref{Fig.13} explains the risk hierarchy in a given system.
Risk Planning is the continuing process of developing an organized, comprehensive approach to risk management. The initial planning includes: establishing a strategy; establishing goals and objectives; planning assessment, handling, and monitoring activities; identifying resources, tasks, and responsibilities; organizing and training risk management IPT members; establishing a method to track risk items; and establishing a method to document and disseminate information on a continuous basis \cite{leonard1999systems}. Figure 14 shows the concept of risk management control and feedback. We can extract information from feedback in cognitive radar using these risk management concepts.  The idea of risk management can especially help us to reduce noise based of estimation error which will be discussed later.
\begin{figure}[h!]
	\includegraphics[width=\linewidth]{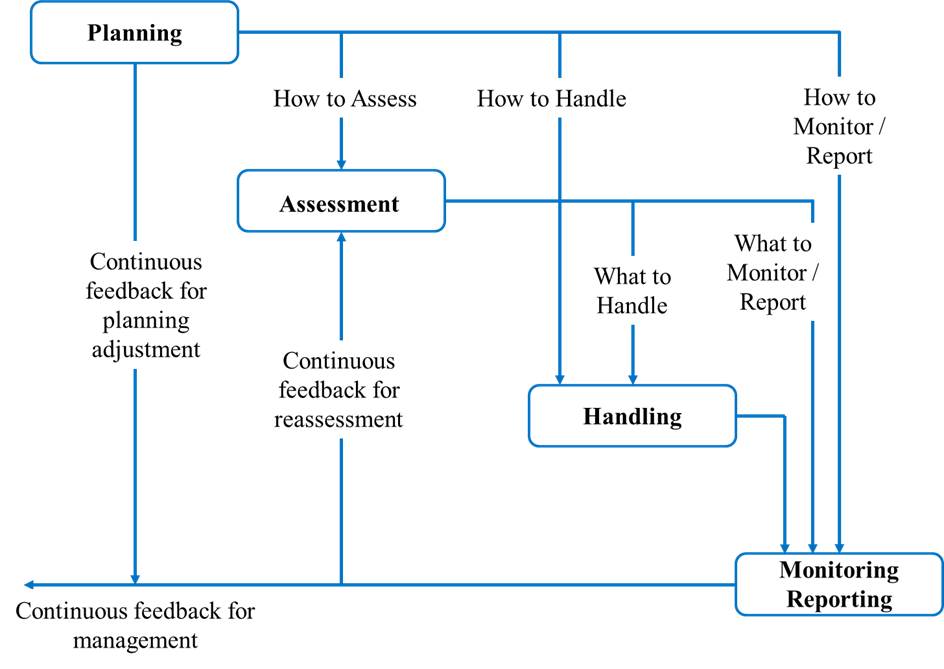}
	\caption{Risk Management Control and Feedback \cite{leonard1999systems}}
	\label{Fig.14}
\end{figure}

\subsection{Spectrum sensing in cognitive radar}

As discussed previously, perception of the environment is the first function of cognition. From the engineering point of view, we need to sense the environment and extract the distinguishing components from the environment. Like cognitive radio, there is potential interest in using spectrum sensing in cognitive radar. Thus, we can consider the spectrum estimation as the discriminant for sensing the environment. Methods for spectrum sensing can be categorized into four methods:
\begin{enumerate}
	\item Power spectrum using the equation:
	\begin{equation}\label{Eq.2}
	\begin{split}
	&E[| \widehat{x}_{T}( \omega )|^{2}]=E[\frac{1}{T} \int_T^0 x^*(t)e^{i \omega t}dt  \int_T^0 x(t')e^{-i\omega t'}dt'] \\
	&=\frac{1}{T} \int_T^0\int_T^0 E[x^{*}(t)x(t')e^{i\omega(t-t')}dtdt'] 
	\end{split}
	\end{equation}
	\item Space-time processing 
	\item Time-frequency analysis 
	\item Spectral line components
\end{enumerate}
In each approach, different algorithms have been developed which have their pros and cons. Within the space-time processing category, blind source separation methods such as singular-value decomposition (SVD), principal component analysis (PCA), and independent component analysis (ICA) are of high interest \cite{sarraf2016functional} \cite{sarraf2016robust}.
In this section, we propose a new approach called Probabilistic Independent Component Analysis (PICA) which will presumably reduce noise based on estimation error in cognitive radar.

\subsection{Probabilistic Independent Component Analysis}
We present an integrated approach to P ICA that allows for non-square mixing in the presence of Gaussian noise. In order to avoid over fitting, we employ objective estimation for the amount of Gaussian noise through Bayesian analysis of the true dimensionality of the data, i.e. the number of activation and non-Gaussian noise sources. This enables us to carry out probabilistic modelling and achieves an asymptotically unique decomposition of the data. It reduces problems of interpretation, as each final independent component is now much more likely to be due to only one process. In this model, other improvements to standard ICA were considered such as temporal pre-whitening and variance normalisation of time series, the latter being particularly useful in the context of dimensionality reduction when weak activation is present \cite{beckmann2004probabilistic} \cite{sarraf2016functional} \cite{grady2016age} \cite{strother2014hierarchy}.
Similar to the square noise-free case, the PICA model is formulated as a generative linear latent variables model. It is characterised by assuming that the p-variate vector of observations is generated from a set of q statistically independent non-Gaussian sources via a linear instantaneous mixing process corrupted by additive Gaussian noise ?(t):
\begin{equation}\label{Eq.3}
x_{i}=As_{i}+ \mu + \eta _{i}
\end{equation}

Here, $x\_i$ denotes the p-dimensional column vector of individual measurements at location i, $s\_i$ denotes the q-dimensional column vector of non-Gaussian source signals contained in the data and $\eta \_i$ denotes Gaussian noise $\eta \_i \sim
 N(0,\sigma^{2} \sum i)$. The vector $\mu$ defines the mean of the observations $x\_i$ where the index i is over the set of all locations V and p×q matrix A is assumed to be non-degenerate. Solving the blind source separation problem requires finding a linear transformation matrix W is a good approximation to the true source signal s such that:
\begin{equation}\label{Eq.4}
 \widehat{x}=Wx 
\end{equation}
The results show \cite{beckmann2004probabilistic}, however, that conditioned on knowing the number of source signals contained in the data and under the assumption that the data are generated, a linear mixture of independent non-Gaussian source signals confounded by Gaussian noise, there is no non-equivalent decomposition into this number of independent non-Gaussian random variables and an associated mixing matrix; the decomposition into independent components is unique, provided we do not attempt to extract more than q source signals from the data.
Maximum Likelihood estimation for PICA model: we are going to keep the parameter ? fixed at its maximum likelihood estimate:
\begin{equation}\label{Eq.5}
 \mu _{ML}=<x_{i}> 
\end{equation}
Assuming zero-mean sources and we will assume that the mean has been removed from the data. The mean can be always be reintroduced after model estimation using: 
\begin{equation}
x_{i}=A(s_{i}+W_{ \mu ML})+ \widetilde{\eta _{i}} 
\end{equation}
where:
\begin{equation}
\widetilde{\eta _{i}}=\eta _{i}+(I-AW)_{\mu ML}
\end{equation}
After several calculations, the following maximum likelihood solutions are achieved depending on knowledge of the latent dimensionality q.
\begin{equation}
\widehat{A}_{ML} = U_{q} ( \triangle _{q}- \sigma ^2I_{q})^{1/2} 
\end{equation}
\begin{equation}
\widehat{S}_{ML}= ( \widehat{A}^{t} \widehat{A})^{-1}  \widehat{A}^{t}x
\end{equation}
\begin{equation}
 \widehat{ \sigma } ^{2}_{ML}= \frac{1}{p-q} \sum_{l=q+1}^p   \lambda _{l}
\end{equation}
Where $U_{q}$ and $\triangle_{q}$ contain the first q eigenvectors and eigenvalues of U and ? and where Q denotes a q×q orthogonal rotation matrix. $S_{ML}$ represents the maximum likelihood source estimates. $\widehat{ \sigma } ^{2}_{ML}$ denotes ML estimate of $\sigma^{2}$.

\subsection{Parallel computing and processing}
\begin{figure}[h!]
	\includegraphics[width=\linewidth]{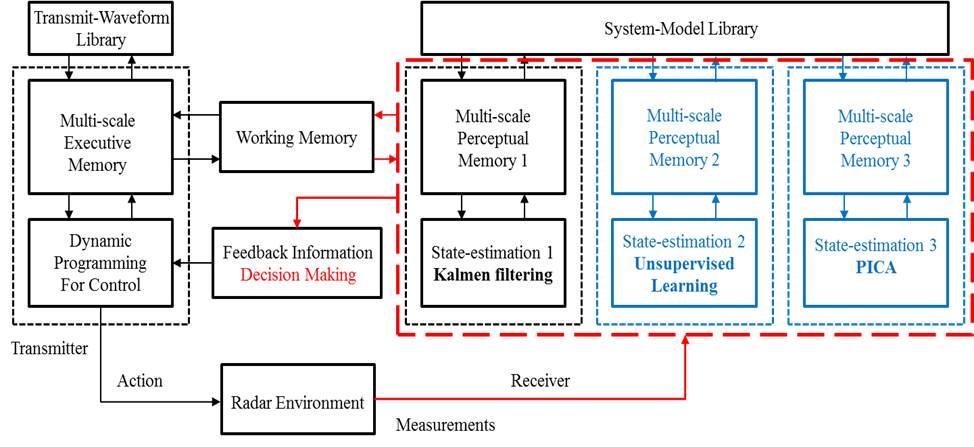}
	\caption{Cognitive radar using parallel computing}
	\label{Fig.15}
\end{figure}

Parallel computing is a concept in computation in which many calculations are carried out simultaneously. The main concept is to divide large problems into smaller ones which can be solved concurrently. There are several different forms of parallel computing such as bit-level, instruction level, data and task parallelism \cite{almasi1989g}. As is known, in cognitive dynamic systems we are faced with difficulties such as: big data processing; information overload and overflow; limited resources; and resource data allocation. In practice, parallel computing and processing are of potential interest to computationally solve above problems. Although it seems that parallel computing will make the system more complex and more expensive hardware will be required, currently available general-purpose graphics processing unit (GPGPU) can be used. As cognitive radar involves real-time systems and online tracking, GPGPU can be used as a part of the system to perform many tasks within the perception-action cycle.
Based on parallel processing concept and the new statistical model proposed previously, the new cognitive radar system shown in Figure 15 is designed. In this system, there are two more processing units than the current design. Also, some changes to the system-model library and feedback information units are required. In state-estimation step, Kalman filtering, unsupervised learning such as deep learning \cite{haykin2012cognitiveB} and PICA are simultaneously executed. The decision making unit is internally added to feedback information block to choose or combine the information received from the state-estimation step. One integrated library is designed which can work simultaneously with perceptual sub-units. The dynamic memory structure is modified in order to communicate with parallel processing unit. The feedback information ? decision making unit processes information coming from parallel processing unit and send processed information to the control unit. In\ref{Fig.15}, the units involving in parallel computing are illustrated in red and two new state-estimation blocks are in blue. The common structure between this proposed design and current cognitive radar structure are in black. Although this new system design must be simulated and implemented on GPGPU, we predict a significant reduction of noise based on estimation error and also a dramatic increase in processing speed.

\section{Report Summary}
We developed and successfully tested our new PySpark-based pipeline on a single node to analyze functional MRI data for extracting brain networks. The new pipeline improved the processing time around 4 times faster than previous works while the accuracy remained at the same value. Furthermore, ease of use, in-memory data processing and storing results in different data structure are some important features of this pipeline. Also, this pipeline can easily expand to several nodes and high performance computing clusters for massive data analysis on large datasets which will definitely improve the processing time and the performance of the pipeline much more than a single node.
\newline
\newline
\textbf{Histroy}
\begin{enumerate}
	\item Origins of the perception action cycle
	\item Fuster?s research on cognition that has influenced cognitive radar
	\item Principals of cognition
\end{enumerate}
\hfill \break
\textbf{Perception-Action Cycle}
\begin{enumerate}
	\item PAC as it applies to cognitive radar
\end{enumerate}
\hfill \break
\textbf{Memory}
\begin{enumerate}
	\item Long-term memory, memory storage, memory retrieval, working memory
	\item Comparison between memory in human cognition and cognitive radar
\end{enumerate}
\hfill \break
\textbf{Attention}

\begin{enumerate}
	\item Comparison of attention in human cognition and cognitive radar
\end{enumerate}
\hfill \break
\hfill \break
\hfill \break
\textbf{Cognitive Radar}
\begin{enumerate}
	\item 	Bayesian filter
	\item	Feedback information
	\item	Dynamic programming 
	\item	State-space model
	
\end{enumerate}
\hfill \break
\textbf{Performance Comparison of Non linear filters in Cognitive Radar (Simulation)}
\begin{enumerate}
	\item CKF, UKF, EKF compared
	\item CKF is better than UKF which is better than EKF in RMSE 
	\item Computation load of EKF is higher than the other two filters	
\end{enumerate}
\hfill \break
\textbf{Risk Management}
\begin{enumerate}
	\item Use concept of risk management to better control parameters especially in noise reduction	
\end{enumerate}
\hfill \break
\textbf{Spectrum Sensing}
\begin{enumerate} 
	\item	Propose new model for space-time processing in spectrum sensing
	\item	New model is PICA 
\end{enumerate}
\hfill \break
\textbf{Parallel Computing and Processing}
\begin{enumerate}
	\item	Simultaneously do complicated calculations or tasks to reduce processing time, increase speed of system or higher precision
	\item	Hardware available (GPGPU)	
\end{enumerate}

\hfill \break

\end{document}